%% file: main.tex
\documentclass{article}

\usepackage{PRIMEarxiv}

\usepackage[utf8]{inputenc} 
\usepackage[T1]{fontenc}    
\usepackage[colorlinks = true,
            linkcolor = green,
            urlcolor  = magenta,
            citecolor = blue,
            anchorcolor = black]{hyperref}
\usepackage{url}            
\usepackage{booktabs}       
\usepackage{amsfonts}       
\usepackage{nicefrac}       
\usepackage{microtype}      
\usepackage{fancyhdr}       
\usepackage{graphicx}       

\usepackage{amsmath}
\usepackage{amsthm}
\usepackage{authblk}

\usepackage{multibib}
\newcites{supp}{Supplementary References}

\usepackage[capitalize]{cleveref}
\crefname{section}{Sec.}{Secs.}
\Crefname{section}{Section}{Sections}
\Crefname{table}{Table}{Tables}
\crefname{table}{Tab.}{Tabs.}

\DeclareMathOperator{\diag}{diag}

\title{Motion-DVAE: Unsupervised learning for fast human motion denoising}

\author[1]{Guénolé Fiche}
\author[1]{Simon Leglaive}
\author[2]{Xavier Alameda-Pineda}
\author[1]{Renaud Séguier}
\affil[1]{CentraleSupélec, IETR UMR CNRS 6164, France}
\affil[2]{Inria, Univ. Grenoble Alpes, CNRS, LJK, France}

\begin{document}
\maketitle

\begin{abstract}
    Pose and motion priors are crucial for recovering realistic and accurate human motion from noisy observations. Substantial progress has been made on pose and shape estimation from images, and recent works showed impressive results using priors to refine frame-wise predictions. However, a lot of motion priors only model transitions between consecutive poses and are used in time-consuming optimization procedures, which is problematic for many applications requiring real-time motion capture. We introduce Motion-DVAE, a motion prior to capture the short-term dependencies of human motion. As part of the dynamical variational autoencoder (DVAE) models family, Motion-DVAE combines the generative capability of VAE models and the temporal modeling of recurrent architectures. Together with Motion-DVAE, we introduce an unsupervised learned denoising method unifying regression- and optimization-based approaches in a single framework for real-time 3D human pose estimation. Experiments show that the proposed approach reaches competitive performance with state-of-the-art methods while being much faster. See the project page at \href{https://g-fiche.github.io/research-pages/motiondvae/}{https://g-fiche.github.io/research-pages/motiondvae/}
\end{abstract}

\input{sec/motion-dvae}


\bibliographystyle{unsrt}  
\bibliography{sample-base}

\setcounter{section}{0}
\setcounter{equation}{0}
\setcounter{figure}{0}
\setcounter{table}{0}
\input{sec/supp}
\bibliographystylesupp{unsrt}  
\bibliographysupp{bib-supp}

\end{document}

%% file: sec/motion-dvae.tex
\section{Introduction}
Human Motion Capture has become a key technology with many applications, such as character animation for the movie and video-game industries~\cite{Avatar, starke2022deepphase, ling2020character}, performance optimization in sports~\cite{swim, groundReaction, AIcoach}, online shopping~\cite{shopping, virtualTrying}, or even preservation of cultural heritage~\cite{cultureHeritage, cyprusDance}. With the rise of AR/VR technologies and the development of the metaverse~\cite{duan2021metaverse, lee2021all}, we will need more accurate and realistic motion capture systems usable by everyone. Although giving good results, traditional maker-based systems can hardly be used on a large scale because they need time-consuming set-up and calibration. Recent advances have enabled motion capture from RGB images and videos~\cite{ExPose:2020, hmrKanazawa17, kocabas2019vibe}. This is a very promising solution to generalize the use of motion capture because it only requires a camera. However, current methods often produce unrealistic motions, especially in the case of occlusions or poor-quality images. The estimated noisy motion capture data needs post-processing before using it for real applications like motion recognition~\cite{petrovich21actor, varol17_surreal} or scene understanding~\cite{holistic++, PSI2019}.

There are numerous motion data post-processing methods, including temporal filtering~\cite{ExploitingTemporalContext}, physical constraints~\cite{walkPhys}, or statistical human motion priors~\cite{StochTrack, trackingCyclic}. More recently, deep learning-based methods gained popularity with the pose priors like VPoser~\cite{SMPL-X:2019}, or Pose-NDF~\cite{tiwari22posendf}, and motion priors like HuMoR~\cite{rempe2021humor} showing impressive results. Generative models are particularly attractive because they can be used in numerous applications, such as motion infilling or movement prediction. Despite impressive results, current pose and motion priors present several issues.
First, many generative motion priors learn transitions between consecutive poses. These models are attractive because they have an excellent physical interpretation and generate realistic motions. However, they are subject to error accumulation in reconstruction, and motion dynamics are usually observable along multiple frames. Moreover, many pose and motion priors are used in optimization procedures, which are often prohibitively time-consuming. This is particularly problematic for multimedia applications like virtual and augmented environments, which require real-time motion capture of users~\cite{WOC}. Finally, very few methods model the noise in the observations.

To address these issues, we introduce Motion-DVAE, a generative motion prior using a dynamical variational autoencoder (DVAE)~\cite{Girin_2021} to capture the short-term dynamics of human motion. While being a generative model like all VAEs~\cite{vaeKingma, rezende2014stochastic}, a DVAE introduces temporal relations among observed and latent variables using recurrent neural networks (RNNs). RNNs can accurately represent the dynamics of human motion, which cannot be summarized as a sequence of independent transitions. 

Together with Motion-DVAE, we propose an unsupervised learned denoising method to refine 3D pose predictions in real time by using the proposed model in regression- or optimization-based procedures. For the learning step, the model parameters are estimated by maximizing the likelihood of the observations through an unsupervised optimization-based procedure. Contrary to prior works, we refine the Motion-DVAE encoder weights instead of the motion latent representation, enabling us to generalize to previously unseen data. Then, once learned on a database representative of the conditions in the test phase, denoising can be performed as a regression task (regression mode). We can also use the unsupervised optimization procedure for per-instance or per-dataset optimization (optimization mode). 

In summary, our contributions are: (i) A generative motion prior, Motion-DVAE, representing short-term dependencies of human motion. (ii) A flexible unsupervised learning framework for real-time human motion denoising in regression- and/or optimization-based procedures. (iii) The proposed method is faster than state-of-the-art (SOTA) methods while showing competitive performance on pose estimation and motion denoising.

\section{Related work}
One of the main applications of the introduced method is to perform 3D human motion estimation from monocular videos. Therefore, we will review prior human pose and shape estimation works. We will then complete our study by reviewing works using human motion priors.

\subsection{Human mesh estimation}
Recent works have made significant progress in 3D human pose and motion estimation~\cite{hmrKanazawa17, SMPLify, SPIN, ExPose:2020, rempe2021humor}. While earlier methods only focused on 3D joint locations~\cite{martinez_2017_3dbaseline, Fuse2d3d, zhou2016sparseness, OcclusionAware, Wandt2019RepNet}, more and more works are model-based. This means that instead of working directly on joint coordinates, they estimate the parameters of a parametric model of the human body~\cite{SCAPE, SMPL:2015, SMPL-X:2019, STAR:2020, xu2020ghum, SUPR}. Here we focus on model-based methods that estimate human pose and shape.

\paragraph{Optimization methods}
Optimization methods iteratively estimate the parameters of a body model given an image or a video. Most of the time, the optimization objective ensures that the projection of 3D predictions is consistent with a set of 2D cues. The first fully automatized method, SMPLify~\cite{SMPLify}, fitted SMPL~\cite{SMPL:2015} to a set of 2D keypoints obtained with an off-the-shelf model~\cite{openpose}. SMPLify uses a set of pose and shape priors to guide the optimization so it converges toward realistic human bodies. Other works use different types of 2D information such as body silhouettes or part segmentation ~\cite{UniteThePeople, holopose, RevitalizingOptimisation}. However, most contributions use 2D keypoints and focus on improving pose and shape priors~\cite{SMPL-X:2019, kolotouros2021prohmr, tiwari22posendf}, as well as the optimization process~\cite{corona2022lvd, song2020human, RevitalizingOptimisation}. 
Various methods estimate motion by taking a sequence of observations as input. In that case, the optimization takes into account temporal motion priors~\cite{ExploitingTemporalContext, CHOMP, rempe2021humor}, and some works jointly estimate moving camera position~\cite{yuan2022glamr, decoupling}. Optimization methods usually give the most accurate results, but the inference is typically much slower than regression methods.

\paragraph{Regression methods}
In most cases, regression methods use a neural network to regress the parameters of a human body model, given an image as input. The canonical example of HMR~\cite{hmrKanazawa17} was the basis for several improvements~\cite{SPIN, ExPose:2020, Kocabas_PARE_2021, zhang2019danet}. Recent advances took advantage of probabilistic modeling to propose multiple mesh predictions given an image~\cite{kolotouros2021prohmr, HierarchicalProb}. Improvements also include the development of video-based approaches. VIBE~\cite{kocabas2019vibe} and other methods~\cite{MEVA, humanMotionKanazawa19, hu2021conditional} estimate motion directly from videos, while~\cite{AttentionRefinement, petrovich21actor} denoise predictions made by an off-the-shelf 3D prediction model with a single forward pass. The main problem for supervised regression methods is the lack of RGB data with 3D annotations, which are very hard to obtain. While~\cite{varol17_surreal, cai2021playing, HierarchicalProb} use synthetic data, another approach is to mix optimization and regression frameworks to learn in an unsupervised fashion. While HMR~\cite{hmrKanazawa17} uses a neural network to replace the optimization step, SPIN~\cite{SPIN} initializes SMPLify~\cite{SMPLify} with a neural network and takes the output of the optimization as supervision.

In this work, we propose a unified framework that can be used in a regression- or optimization-based procedure. We will train Motion-DVAE on noisy data with an unsupervised optimization-based method and use it in regression or optimization modes on previously unseen data. This approach can be compared to the learning process of SPIN~\cite{SPIN} as optimization is used to train a regression method. However, the proposed method is different since optimization is performed directly on the neural network's weights instead of taking optimization results as pseudo-ground-truth for network training. Other works finetuned neural networks for per-instance optimization~\cite{alldieck2019learning, joo2021exemplar}, however to our knowledge, our approach is the first using unsupervised neural network training for per-dataset motion refinement.

\subsection{Pose and motion priors}
Pose and motion priors are crucial for accurate and realistic motion capture from noisy observations. The first proposed approaches relied on statistical models such as principal component analysis~\cite{StochTrack,trackingCyclic} or Gaussian mixture models learned on pose data~\cite{SMPLify}, and physical constraints~\cite{walkPhys}. With the rise of deep learning, new approaches emerged, using physics of motion~\cite{neuralMoCon, yuan2021simpoe, PhysCapTOG2020}, and adversarial training~\cite{hmrKanazawa17, HPGan, lyu2021learning}.
Recent pose~\cite{SMPL-X:2019} and motion~\cite{rempe2021humor, petrovich21actor, mvae, Habibie2017ARV, UnifiedMotionSynthesis, TaskGenericVAE, bie2022hit, ling2020character, action2motion} priors using VAEs~\cite{vaeKingma, rezende2014stochastic} showed very impressive results. In particular, ACTOR~\cite{petrovich21actor} learns an action-conditioned VAE-transformer prior, and HuMoR~\cite{rempe2021humor} uses a conditional VAE (CVAE) to learn transitions between consecutive poses. More recently, Pose-NDF~\cite{tiwari22posendf} outperformed SOTA models~\cite{rempe2021humor, SMPL-X:2019} for pose and motion denoising by learning a manifold of human poses.

In this work, we use a VAE-based model for motion-prior learning in line with recent works. RNNs used in DVAE enable us to model motion short-term dependencies efficiently.

\section{Motion-DVAE}
\paragraph{Notation.} In all the following, we will use the notation $\mathcal{N}(\mu,\sigma)$ to denote a Gaussian distribution with mean $\mu$ and a diagonal covariance matrix with diagonal $\sigma$.

\subsection{State representation}
We want to represent the state of a moving person. A motion can be seen as a sequence of poses leading to translations and rotations in space. Thus, we take $r \in \mathbb{R}^3$ the global translation, $\phi \in \mathbb{R}^3$ the global orientation, and $\Theta \in \mathbb{R}^{21*3}$ the pose in axis-angle representation. Those parameters can be fed to the SMPL model~\cite{SMPL:2015} to generate a human body mesh, given body shape parameters $\beta \in \mathbb{R}^{10}$. It will output a vertex mesh $V \in \mathbb{R}^{6890\times3}$ and a set of joints $J \in \mathbb{R}^{22\times3}$. 
In practice, we notice that giving the model parameters' velocities helps the DVAE to reconstruct human motions. Therefore, we also take $\dot{r}$, $\dot{\phi}$ and $\dot{\Theta}$ to represent the state. In the end, a moving person can be represented as a vector:
\begin{equation}
    x = [r, \dot{r}, \phi, \dot{\phi}, \Theta, \dot{\Theta}] \in \mathbb{R}^{46*3}.
\end{equation}
We can obtain the meshes and joints associated with $x$ using the function:
\begin{equation}
    V, J = \mathcal{M}_\beta(x),
\end{equation}
which corresponds to the SMPL forward pass computed from the appropriate entries of $x$ and $\beta$.

\subsection{Motion-DVAE model} \label{DVAE_model}
We model human motion as a causal process: each state depends only on the past. We propose a DVAE \cite{Girin_2021} with a sequence of latent vectors $z_{1:T}$ to implicitly capture the temporal dependencies of human motion.

\paragraph{Generative model} The joint distribution of the introduced Motion-DVAE is defined as:
\begin{equation}\label{genModel}
    p_\theta(x_{1:T}, z_{1:T} | x_0) = p_{\theta_z}(z_1 | x_0) p_{\theta_x}(x_1 | z_1, x_0) \times \prod_{t=2}^T p_{\theta_z}(z_t | z_{1:t-1},x_0) p_{\theta_x}(x_t | z_{1:t}, x_0),
\end{equation}
where $\theta_x$ and $\theta_z$ represent the decoder and the prior network parameters and $\theta = \theta_z \cup \theta_x$. From the DVAE perspective, $x_0$ is fixed and can be considered a parameter. Indeed we model the dynamics of motion, and $x_0$ is just the starting point of it. The generative model of the DVAE is set as follows:
\begin{equation}
    p_{\theta_x}(x_t | z_{1:t}, x_0) = \mathcal{N}(x_t;\mu_{\theta_x}(z_{1:t}, x_0),Id),
\end{equation}
where $\mu_{\theta_x}(z_{1:t}, x_0)$ is the output of the decoder network. We can interpret $z_{1:t}$ as a sequence of transitions allowing us to go from $x_0$ to $x_t$. $z_t$ should then depend on $x_0$ and previous transitions.  Hence, we assume $z_{1:0} = \emptyset$ and define the prior model as follows:
\begin{equation}
    p_{\theta_z}(z_t | z_{1:t-1}, x_0) = \mathcal{N}(z_t;\mu_{\theta_z}(z_{1:t-1}, x_0),\sigma_{\theta_z}(z_{1:t-1}, x_0)).
\end{equation}
The generative model will be fixed during denoising learning. Implementation details are given in \cref{subsec:implementation_gen}.

One can notice that given the latent sequence $z_{1:t}$, the DVAE does not use past states $x_{1:t-1}$ to predict $x_t$. Thus, contrary to most motion models like \cite{rempe2021humor}, we do not need to alternate between sampling $x_t$ and $z_t$ to sample a motion sequence. Furthermore, models that use previous predictions $x_{1:t-1}$ for predicting $x_t$ are more likely to be subject to posterior collapse~\cite{dai2018diagnosing, UsualSuspects, lucas2019don}.

\paragraph{Inference model} From the joint distribution \cref{genModel}, D-separation~\cite{Girin_2021} yields the following approximate posterior distribution:
\begin{equation}\label{infModel}
    q_\phi(z_{1:T}|x_{0:T}) = q_\phi(z_0|x_{0:T})\prod_t q_\phi(z_t | z_{1:t-1}, x_{t:T}, x_0).
\end{equation}
We define the inference model as follows:
\begin{equation}
    q_\phi(z_t | z_{1:t-1}, x_{t:T}, x_0) = \mathcal{N}(z_t;\mu_{\phi}(z_{1:t-1}, x_{t:T}, x_0),\sigma_{\phi}(z_{1:t-1}, x_{t:T}, x_0)),
\end{equation}
where $\mu_\phi$ and $\sigma_\phi$ are the output of the encoder network and $z_{1:0} = \emptyset$.

Implementation details of the inference model are given in \cref{implementation_inf}.

\paragraph{Loss functions} The loss is computed using the evidence lower bound (ELBO)~\cite{Girin_2021} derived from the inference and generative models. We also add a regularization term at the output of the SMPL model to enforce the final human mesh to be as close as possible to the original mesh, consisting of a squared reconstruction error on meshes and joints. More information about loss functions is available in \cref{loss_dvae}, and learning settings are detailed in \cref{dvae_settings}.

\section{Unsupervised learned denoising}
In this section, we propose to exploit Motion-DVAE to refine the pose estimate in the context of monocular videos or to recover motion from noisy 3D data. We introduce a robust noise model which can be tuned depending on the use case.

\subsection{Problem definition and modeling}
Suppose that we have 3D noisy observations in SMPL format:
\begin{equation}\label{observations}
    y_t^{raw} = [r_t, \phi_t, \Theta_t, \beta_t].
\end{equation}
Those parameters' velocities can be obtained through finite differences in a deterministic way, leading to $y_t = [r_t, \dot{r}_t, \phi_t, \dot{\phi}_t, \Theta_t, \dot{\Theta}_t]$.
We take as observations the SMPL-H~\cite{MANO} body joints $J_t \in \mathbb{R}^{22\times3}$ and a subset of mesh vertices $V_t \in \mathbb{R}^{43\times3}$ which can be seen as virtual motion capture markers~\cite{MoSh}:
\begin{equation}
    y^{3d}_t = [J_t, V_t] \in \mathbb{R}^{65*3}.
\end{equation}
From the 3D noisy observations, our goal is to infer the initial state $x_0$, the sequence of latent variables $z_{1:T}$, and the body shape $\beta$ that correspond to the ground-truth clean motion.

\paragraph{Obtaining 3D observations from an RGB video.} As stated in \cref{observations}, the proposed unsupervised learned denoising works on 3D global human motion noisy data. In order to obtain noisy motion capture data from RGB videos, we propose SPIN-t, a procedure built on SPIN's~\cite{SPIN} frame-wise local pose estimations (in frame coordinates) to obtain global motion (in world coordinates). Details about SPIN-t are available in \cref{sec:spint}.

\paragraph{Joint distribution.} We define the generative model of the noisy observations with the following joint distribution:
\begin{equation}
    p_\theta(y_{0:T},v_{0:T},z_{1:T} | \beta, x_0) = p(v_0)p(y_0|x_0,\beta,v_0) \times \prod_{t=1}^T p_\theta(z_t|z_{1:t-1},x_0) p(v_t) p_\theta(y_t|z_{1:t},\beta, v_t, x_0),
\end{equation}
with:

$\triangleright$ $p(y_0|x_0,\beta,v_0) = \mathcal{N}(y^{3d}_0; \mathcal{M}_\beta(x_0),v_0)$. This likelihood distribution is equivalent to considering that $y_0$ is equal to the output of the SMPL model applied on the initial state $x_0$ plus a heteroscedastic Gaussian noise with zero mean and variance coefficients $v_0$. Note that the distribution is defined on the 3D predictions instead of SMPL parameters. This is more convenient since we do not need to define a probabilistic distribution over $\mathcal{SO}^3$~\cite{HierarchicalProb}.

$\triangleright$ $p_\theta(y_t|z_{1:t},\beta,v_t, x_0) = \mathcal{N}(y^{3d}_t; \mathcal{M}_\beta(\mu_{\theta_x}(z_{1:t},x_0)),v_t)$. This likelihood distribution is very similar to the previous one, except that we will use the Motion-DVAE decoder $\mu_{\theta_x}$ to parameterize $x_t$ by $x_0$ and a sequence of latent variables $z_{1:t}$.

$\triangleright$ $p_\theta(z_t |z_{1:t-1}, x_0) = \mathcal{N}(z_t; \mu_{\theta_z}(z_{1:t-1}, x_0), \sigma_{\theta_z}(z_{1:t-1}, x_0))$. Here, we feed the initial state and past latent variables to the Motion-DVAE prior network $(\mu_{\theta_z}, \sigma_{\theta_z})$ to obtain the current latent variable $z_t$. We use the same prior distribution for $p_\theta(z_1 |x_0)$ by assuming $z_{1:0} = \emptyset$.

$\triangleright$ $p(v_{t}) = \prod_{j,d}p(v_{t,j,d})$ where $j \in \{1:22\}$ stands for the joint index, $d \in \{1:3\}$ is the spatial dimension ($x$, $y$, or $z$) and $p(v_{t,j,d}) =  \mathcal{IG}\left(v_{t,j,d};\frac{\lambda}{2},\frac{\lambda}{2}\right)$. This is a confidence model defined on 3D observations. $\mathcal{IG}$ stands for the Inverse-Gamma distribution, the larger $\lambda$ the higher confidence in observations. This variance model, along with the above-defined likelihood model $p_\theta(y_t|z_{1:t},\beta,v_t, x_0)$ is equivalent to assuming that the 3D observations' positional error follows a Student-t distribution, which is more robust and adaptable than the Gaussian distribution~\cite{robustFrontalization}.

\paragraph{Approximate Posterior} We derive the approximate posterior of the latent variables given the noisy observations from the previous joint distribution and D-separation~\cite{geiger1990identifying, bishop2006pattern} (see \cref{posterior}):
\begin{equation}\label{appPost}
    q_{\phi,\gamma,\omega}(x_0,\beta,z_{1:T},v_{0:T}|y_{0:T}) = q(\beta|y_{0:T}) q_\omega(x_0 | y_{0:T}) q_\gamma(v_0|y_0,x_0) \times  \prod_{t=1}^T \Bigl[q_\gamma(v_t|z_{1:t},y_t,x_0) q_\phi(z_t|z_{1:t-1},y_{t:T},x_0)\Bigr],
\end{equation}
where:

$\triangleright$ $q(\beta|y_{0:T})$ is modeled as a Dirac delta function centered on the observations' average shape.

$\triangleright$ $q_\phi(z_t|z_{1:t-1},y_{t: T},x_0)$ corresponds to the approximate posterior distribution of the DVAE described in \cref{DVAE_model}. Hence, we will finetune the encoder weights during denoising learning.

$\triangleright$ $q_\omega(x_0|y_{0:T})$ and $q_\gamma(v_t |z_{1:t},y_t,x_0)$ are respectively Gaussian and Inverse Gamma distributions and  will be implemented by two new neural networks (initial state and noise predictors) learned during the denoising learning step.

This inference model will be used to estimate the latent variables from the noisy observations, particularly the initial state $x_0$ and the latent motion $z_{1:T}$, which encode the ground truth motion $x_{1:T}$.

\subsection{Unsupervised denoising learning} \label{learningProcess}
The aim of unsupervised denoising learning is to learn the parameters of the inference model defined in \cref{appPost}. The objective is to push the approximate posterior $q_{\phi,\gamma,\omega}(x_0,\beta,z_{1:T},v_{0:T}|y_{0:T})$ towards the intractable exact posterior distribution $p_\theta(x_0,\beta,z_{1:T},v_{0:T}|y_{0:T})$.
We can show that this is equivalent to maximizing an ELBO~\cite{neal1998view}, yielding the following objective function (see \cref{detailed_elbo}):
\begin{equation}
    \mathcal{L}(\phi,\gamma,\omega) = \mathcal{L}_{rec} + \mathcal{L}_{KL}^{DVAE} + \mathcal{L}_{KL}^{noise},
    \label{lossUnsupervided}
\end{equation}
where
\begin{equation}
    \mathcal{L}_{rec} = - \frac{1}{2}\mathbb{E}_{q_{\phi,\gamma,\omega}}\Biggl[\sum_{j,d} \frac{1}{v_{0,j,d}} (y_{0,j,d}-\mathcal{M}_\beta(x_0)_{j,d})^2 + \sum_{t,j,d} \frac{1}{v_{t,j,d}}(y_{t,j,d} - \mathcal{M}_\beta(\mu_{\theta_x}(x_0, z_{1:t}))_{j,d})^2\Biggr]
\end{equation}
is a negative MSE weighted by the inverse of the estimated noise variance for each observation and
\begin{equation}
    \mathcal{L}_{KL}^{DVAE} = - \sum_{t=1}^T \mathbb{E}_{q_{\omega}q_{\phi}} \Big[ D_{KL}\big(\, q_\phi(z_t|z_{1:t-1},y_{t:T},x_0) \parallel p_\theta(z_t|x_0,z_{1:t-1})\, \big) \Big];
\end{equation}
\begin{equation}
    \mathcal{L}_{KL}^{noise} = - \sum_{t=1}^T \mathbb{E}_{q_{\omega}q_{\phi}}      \Big[D_{KL}\big(q_\gamma(v_t |z_{1:t},y_t,x_0) \parallel p(v_t) \big)\Big]
\end{equation}
push the approximate posteriors towards the priors.

To learn the inference model, we will jointly optimize $\phi$, $\gamma$, and $\omega$, which are the weights of the Motion-DVAE encoder initially learned on AMASS~\cite{AMASS}, noise predictor, and initial state predictor networks. No other parameter will be optimized, so there will not be any mandatory optimization process when performing denoising on new data. This enables dramatic speed-up of denoising new data. Note that Motion-DVAE prior and decoder networks are fixed during the denoising training process. This enables keeping the motion prior unchanged and focusing on filling the gap between noisy observations and clean motion capture data. Implementation details are given in \cref{denoising_implementation}.

\begin{figure}[t]
    \centering
    \includegraphics[width=0.5\textwidth]{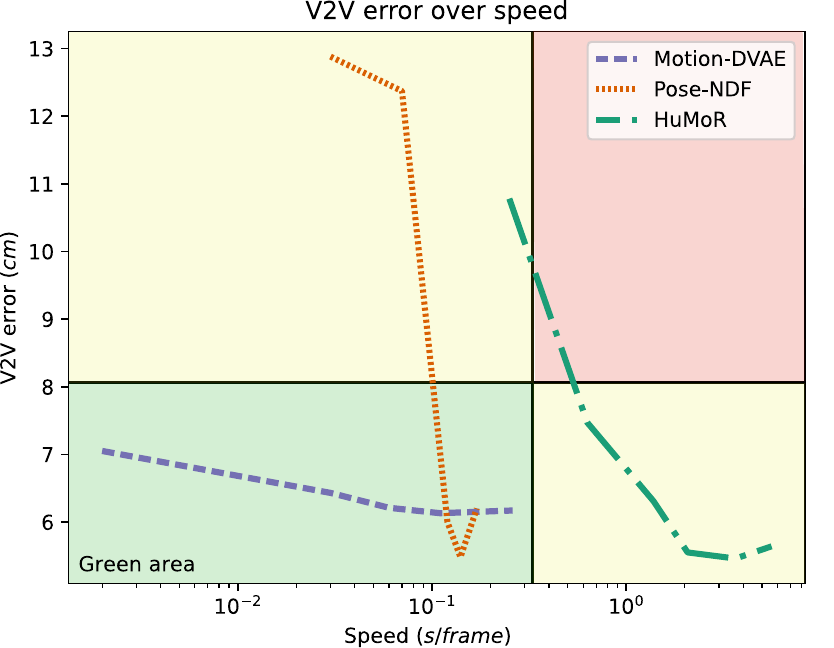}
    \caption{Evolution of the accuracy in V2V versus speed.}
    \label{evolution}
\end{figure}

\subsection{Denoising new data} \label{inferenceProcess}
For denoising new data, the final prediction is the posterior mean estimate of $x^{3d}_{0:T}$. First, we need to predict the initial state:
\begin{align}
    \hat{x}^{3d}_0 = \mathbb{E}_{q_{\omega}(x_0|y_{0:T})}[\mathcal{M}_\beta(x_0)] \simeq \mathcal{M}_\beta(\mu_\omega(y_{0:T})).
\end{align}
where $\mu_\omega$ is the mean vector of $q_\omega(x_0 | y_{0:T})$.
To infer $\hat{x}_0$, we have to feed the initial state predictor with observations and take the mean of the output distribution.
For predicting $x_{1:T}$ we will use:
\begin{align}
    \hat{x}_{1:T} = \mathbb{E}_{q_{\phi,\gamma,\omega}}\Bigl[\mathbb{E}_{p_\theta(x_{1:T}|x_0,\beta,z_{1:T},v_{0:T},y_{0:T})}[x_{1:T}]\Bigr].
\end{align}
This leads to the following estimate:
\begin{equation}
    \hat{x}^{3d}_{t,j,d} = \mathbb{E}_{q_{\phi,\gamma,\omega}}\left[\frac{v_{t,j,d} \mathcal{M}_\beta(\mu_{\theta_x}(x_0,z_{1:t-1})_{j,d})+  y^{3d}_{t,j,d}}{v_{t,j,d} + 1}\right].
    \label{infFormula}
\end{equation}
More detailed calculations are available in \cref{final_pred}. We assume that $\mathcal{M}_\beta(x_t)$ follows a Gaussian distribution with mean $\mathcal{M}(\mu_{\theta_x}(x_0,z_{1:t-1})_{j,d})$ and unit variance. This is an approximation because $x_t$ being a Gaussian distribution does not mean that $\mathcal{M}_\beta(x_t)$ is also Gaussian and the variance of $x_t$ and $\mathcal{M}_\beta(x_t)$ are not equal. However, this approximation is used in state-of-the-art methods~\cite{tiwari22posendf} and works well in practice.

We can intuitively interpret the final clean motion estimate in \cref{infFormula}: if the confidence in observations is very high, $v_{t,j,d}$ will be a small value, and the final predictions will be very close to $y^{3d}_{t,j,d}$. Conversely, if $v_{t,j,d}$ is large, the final prediction will rely more on the Motion-DVAE output. The process to estimate $\hat{x}_{1:T}$ compromises observations and the Motion-DVAE output weighted by the estimated variance. It needs recursive Monte Carlo sampling on $x_0$ and $z_{1:T}$ to calculate the expectation of the noise for the initial state and the motion. Despite this necessary sampling, denoising can be made in real time because each sample only consists of a forward pass in a neural network, which is almost instantaneous in our case.

\section{Experimental results}
This section evaluates the proposed unsupervised learned denoising framework with Motion-DVAE for human motion data denoising. Experiments on RGB videos and an ablation study are available as supplementary material (see \cref{sec:additional_exp}). We evaluate the proposed method for each experiment and compare it to the state-of-the-art in terms of accuracy and speed. All tests have been run on a single Nvidia GeForce GTX 1070.

For experiments on human motion data denoising we use AMASS~\cite{AMASS}. The AMASS dataset~\cite{AMASS} is a large database of human motion that unifies several motion capture databases by representing them in SMPL format. It contains more than 300 subjects and 10k motions, covering various shapes and poses. We sub-sample AMASS to 30Hz and follow the recommended training, validation, and testing splits.

\subsection{Denoising noisy motion capture data}
\begin{table}[!ht]
    \centering
    \caption{Motion denoising on noisy AMASS: We compare our method (bottom part) and SOTA (middle part) performance for 3 different noise values.}
    \begin{tabular}{c|c|ccc|ccc}
                                           & Speed $\downarrow$ & \multicolumn{3}{c|}{V2V $\downarrow$}                     & \multicolumn{3}{c}{Acceleration} \\
        \midrule
        Ground truth                        & -                  & \multicolumn{3}{c|}{-}                                    & \multicolumn{3}{c}{6.32}                      \\
        Observations                       & -                  & 8.68              & 12.89             & 16.93             & 137.44            & 202.56            & 264.28        \\
        \hline
        HuMoR~\cite{rempe2021humor}        & 5.60               & \underline{5.61}  & \textbf{6.31}     & \underline{7.97}  & 4.36              & \textbf{5.55}     & \underline{8.76} \\
        Pose-NDF~\cite{tiwari22posendf}    & 0.17               & 5.89              & 6.90              & \textbf{7.89}     & 9.64              & 10.03             & 10.84         \\
        \hline
        Regression (ours)                  & \textbf{0.001}     & 6.25              & 7.69              & 9.36              & \textbf{7.64}     & \underline{7.99}  & \textbf{8.57} \\
        Optimization (ours)                & \underline{0.06}   & \textbf{5.02}     & \underline{6.74}  & 8.72              & \underline{8.09}  & 9.98              & 9.49 
    \end{tabular}
    \label{noisyAmass}
\end{table}

First, we evaluate the unsupervised learned denoising on noisy motion capture data. To generate noisy mocap data, we use the same procedure as Pose-NDF~\cite{tiwari22posendf}: we add a Gaussian noise to the rotation of each joint in the axis-angle representation. We test our method on sequences of 2 seconds. We set the noise standard deviation to 0.1, 0.15, and 0.2, corresponding to initial per-vertex errors of 8.68, 12.89, and 16.93~cm. The observations are the noisy vertices (virtual Mocap markers) and joints, and we initialize the denoising with the corresponding noisy SMPL parameters.

Starting from the proposed Motion-DVAE trained on AMASS, we use the unsupervised learning process described in \cref{learningProcess} on the noisy AMASS training data. We then use our model in both regression and optimization modes. The inference model is fixed for regression mode, and predictions are made through a single forward pass in Motion-DVAE. In optimization mode, the inference model is optimized on test samples in an unsupervised fashion for a fixed number of iterations using \cref{learningProcess}. Note that all optimization methods, such as HuMoR~\cite{rempe2021humor} and Pose-NDF~\cite{tiwari22posendf}, perform optimization on the test set. The difference between the proposed approach and those methods is that we can use Motion-DVAE in regression mode and that optimization variables are neural networks parameter, enabling us to optimize the inference model for the whole test set at once independently of the number of sequences.

Instead of using \cref{infFormula} for denoising, we use the initial state predictor and Motion-DVAE output as the prediction. This saves us from learning to predict the noise on every 6890 vertices, which would be complicated in practice. Then, using the proposed Student-t noise model for training and optimization would not make as much sense since we do not use it for final predictions. Thus, we use very high values for noise parameter $\lambda$, which is equivalent to choosing a Gaussian prior distribution for the noise. Adding a Gaussian noise to the SMPL pose parameter does not give a Gaussian noise in 3D observations; however, this approximation was also made in the SOTA approach~\cite{tiwari22posendf} and works well in practice.
We compare our results in regression and optimization modes with public implementations of SOTA methods~\cite{tiwari22posendf,rempe2021humor} in \cref{noisyAmass}, regarding speed (in $s/frame$), per-vertex-error (V2V in $cm$), and joints acceleration (in $m/s^2$). Joints acceleration measures the motion plausibility, the best joint acceleration is the closest to the ground truth value (6.32).

The proposed method is much faster than the SOTA methods. In the regression mode, the proposed method is 170 times faster than Pose-NDF and 5000 times faster than HuMoR. Regarding per-vertex error, the proposed approach in optimization mode outperforms the SOTA methods for an introduced noise of 8.68 cm. Our performance is similar to the SOTA models for an added noise of 12.89 cm in observations. However, the introduced method is less efficient than others for an added noise of 16.93 cm. This performance degradation for our approach when the added noise is large is unsurprising. Indeed, we never optimize the inputs of Motion-DVAE, which are always noisy observations. When the introduced noise is significant, correcting the error through a single forward pass becomes hard. In terms of joint acceleration, HuMoR produces the most realistic motions. However, it sometimes produces over-smoothed predictions, and the mean joint acceleration highly depends on the noise value. Pose-NDF produces unnatural motions according to the joint acceleration criteria, which is unsurprising since its only temporal model is a smoothing term over time. Our method produces reasonably smooth motions and is not very sensitive to the noise value. One can notice that optimization decreases the per-vertex error but also increases joint acceleration. This observation is expected: when Motion-DVAE is trained on an extensive training set and used in regression mode, it tends to produce smooth motions. In the optimization mode, the proposed model is adapted to specific noisy motion samples, which improves the prediction in terms of per-vertex error at the expense of smoothness.

\begin{figure}[t]
    \centering
    \includegraphics[width=0.7\textwidth]{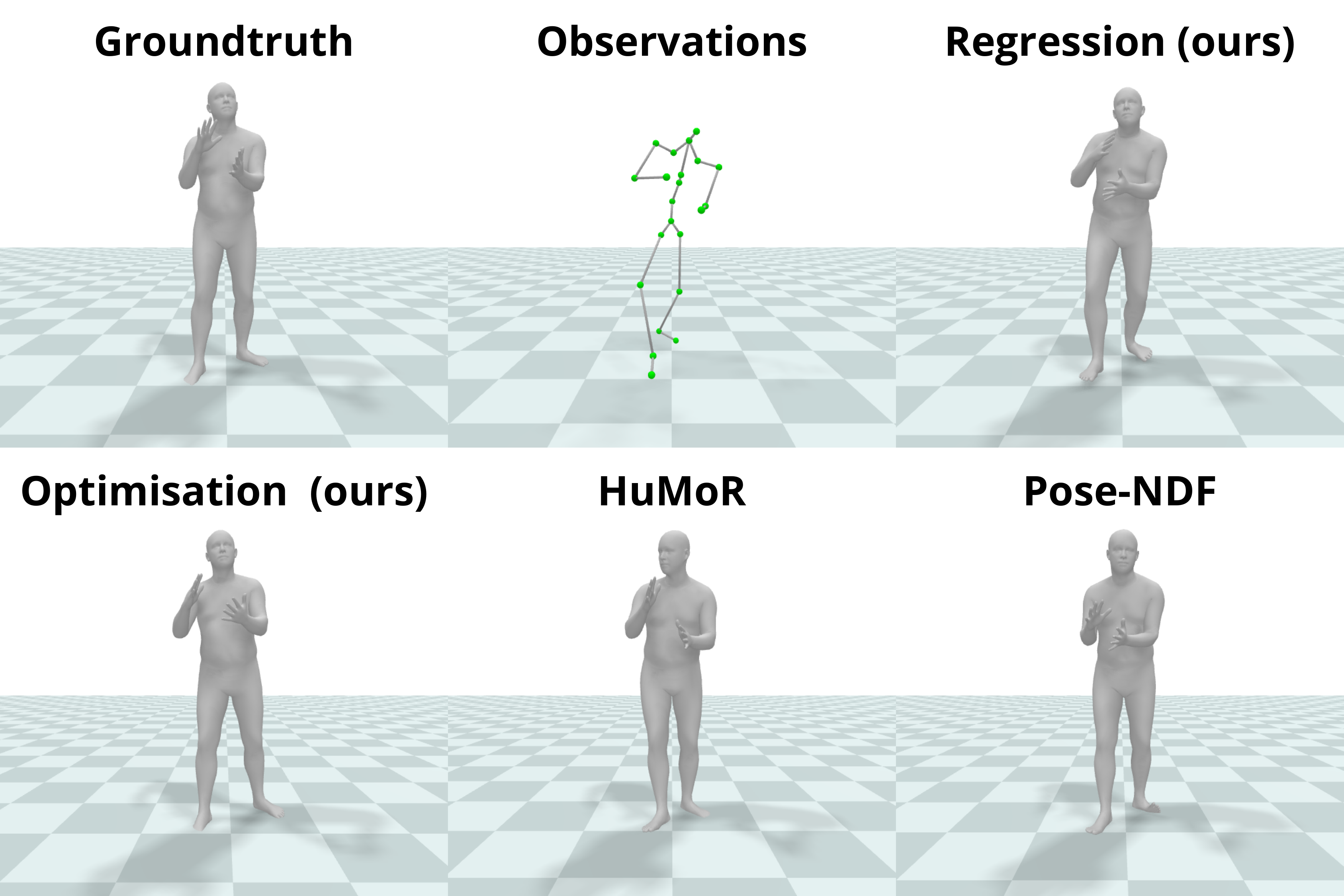}
    \caption{Qualitative results for motion denoising.}
    \label{qualAmass}
\end{figure}

Qualitative results are shown on \cref{qualAmass}. For Motion-DVAE, optimization was performed on the whole noisy AMASS test set, not only on the visualized sequence. Qualitative results on videos are available on the project website.

\subsection{Methods accuracy versus speed}
The previous experiment fixed the number of iterations for the proposed and state-of-the-art optimization-based methods. The proposed inference model was optimized for 50 iterations. We used the implementation default number of iterations for \cite{rempe2021humor} and \cite{tiwari22posendf}. We are now interested in investigating the methods' accuracy as a function of the speed. For this purpose, we vary the number of iterations. For the proposed method, we use the regression mode (0 optimization iterations) as well as the optimization mode for a number of steps between 20 and 250. The methodology for reducing \cite{rempe2021humor} and \cite{tiwari22posendf} optimization processes as well as experiments on the output plausibility as a function of speed are exposed in the supplementary material (see \cref{sec:additional_exp}).

Experiments were performed on a randomly selected subset of 54 sequences from the test set of noisy AMASS, with a noise standard deviation of 0.15. We arbitrarily define acceptability thresholds for each metric, these could differ depending on the use case. For the speed, we consider that a method should not be more than 10 times slower that the frame rate. Beyond that, we consider that the method becomes difficult to use on a large scale. For accuracy, we expect the method to decrease the error of the noisy observations by at least 33 percent.

Results are shown in \cref{evolution}. HuMoR obtains good accuracy performance, but its slowness makes it unusable in practice. Pose-NDF with 5 steps satisfies all the criteria, however Motion-DVAE is the best choice to obtain good accuracy in a reasonable amount of time.

\section{Conclusion}
We introduced Motion-DVAE, a human motion prior based on the dynamical variational autoencoder framework for modeling short-term dependencies of human movement. Together with Motion-DVAE, we proposed an unsupervised learned denoising method allowing for robust noise modeling. This procedure can be used for motion denoising in regression- or optimization-based frameworks and adapts to new data in real time. Our method is more than 100 times faster than SOTA approaches for motion denoising while showing similar accuracy and plausibility. It is state-of-the-art for local motion estimation from videos, while being about 50 times faster than SOTA optimization methods.

Future work includes modeling environmental interactions, such as ground and object contact. This would probably help improve the global predictions, which is one of the weaknesses of Motion-DVAE. Another potential development would be improving noise modeling, for example, by learning visibility.

\section{Acknowledgements}
This study is part of the EUR DIGISPORT project supported by the ANR within the framework of the PIA France 2030 (ANR-18-EURE-0022). This work was performed using HPC resources from the “Mésocentre” computing center of CentraleSupélec, École Normale Supérieure Paris-Saclay, and Université Paris-Saclay supported by CNRS and Région Île-de-France.

%% file: sec/supp.tex
\onecolumn

\appendix

\begin{center}
{\huge \textbf{Supplementary material}}    
\end{center}

\vspace{10 pt}

\section{Motion-DVAE training}
\subsection{Implementation of the generative model}\label{subsec:implementation_gen}
We recall the generative model of Motion-DVAE:
\begin{equation}
    p_\theta(x_{1:T}, z_{1:T} | x_0) = p_\theta(z_1 | x_0) p_\theta(x_1 | z_1, x_0) \prod_{t=2}^T p_\theta(z_t | z_{1:t-1},x_0) p_\theta(x_t | z_{1:t}, x_0),
\end{equation}

We implement $p_\theta(x_t | z_{1:t}, x_0) = \mathcal{N}(x_t;\mu_{\theta_x}(z_{1:t}, x_0),Id)$ with:
\begin{itemize}
    \item $h_0^x = d_{ini}(x_0)$,
    \item $h_t^x = d_{h^x}(h_{t-1}^x, z_t)$,
    \item $\mu_{\theta_x}(h_t^x)=d_x(h_t^x)$.
\end{itemize}

For $p_\theta(z_t | z_{1:t-1}, x_0) = \mathcal{N}(z_t;\mu_{\theta_z}(z_{1:t-1}, x_0),\sigma_{\theta_z}(z_{1:t-1}, x_0)),$ we take:
\begin{itemize}
    \item $h_0^z = d_{ini}(x_0)$
    \item $ h_t^z = d_{h^z}(h_{t-1}^z, z_t)$
    \item $[\mu_{\theta_z}(h_t^z),\sigma_{\theta_z}(h_t^z)]=d_z(h_t^z)$
\end{itemize}

$d_{h^x}$ and $d_{h^z}$ are recurrent neural networks and are both implemented with the same LSTM. $d_{ini}$, $d_x$, and $d_z$ are MLPs with 2 hidden layers using group normalization with 16 groups and ReLU activations. The latent dimension is 48, while the LSTM hidden state dimension is 1024.

\subsection{Implementation of the approximate posterior}\label{implementation_inf}
The inference model is:
\begin{equation}
    q_\phi(z_{1:T}|x_{0:T}) = q_\phi(z_0|x_{0:T})\prod_t q_\phi(z_t | z_{1:t-1}, x_{t:T}, x_0).
\end{equation}

$q_\phi(z_t | z_{1:t-1}, x_{t:T}, x_0) = \mathcal{N}(z_t;\mu_{\phi}(z_{1:t-1}, x_{t:T}, x_0),\sigma_{\phi}(z_{1:t-1}, x_{t:T}, x_0))$ is implemented by:
\begin{itemize}
    \item $h_t^{glob} = [h_{t-1}^{lat},h_t^{data}]$,
    \item $h_0^{lat} = e_{ini}(x_0)$,
    \item $h_t^{lat} = e_h^{lat}(h_{t-1}^{lat}, z_t)$,
    \item $h_0^{data} = 0$,
    \item $h_t^{data} = e_h^{data}(h_{t+1}^{data}, x_t)$,
    \item $[\mu_{\phi}(h_t^{glob}),\sigma_{\phi}(h_t^{glob})]=e_{glob}(h_t^{glob})$.
\end{itemize}

$e_h^{lat}$ and $e_h^{data}$ are LSTM neural networks, while $e_{ini}$ and $e_{glob}$ are MLPs. Implementations of LSTM and MLP are similar to the generative model.

\subsection{Loss function}\label{loss_dvae}
From \cref{genModel} and \cref{infModel}, following~\citesupp{Girin_2021supp}, the ELBO of Motion-DVAE is:
\begin{equation}
    \mathcal{L}(\theta,\phi;x_{0:T}) = \sum_{t=1}^T \mathbb{E}_{q_\phi}[\log p_\theta(x_t | z_{1:t}, x_0)] - \sum_{t=1}^T \mathbb{E}_{q_\phi}[D_{KL}(q_\phi(z_t|z_{1:t-1},x_{t:T},x_0) || p_\theta(z_t|z_{1:t-1}, x_0))].
\end{equation}

We define the loss function of Motion-DVAE:
\begin{equation}
    \mathcal{L}^{train}(\theta,\phi;x_{0:T}) = \mathcal{L}_{rec}(\theta,\phi;x_{0:T}) + \mathcal{L}_{KL}(\theta,\phi;x_{0:T}) + \mathcal{L}_{reg}(\theta,\phi;x_{0:T}).
\end{equation}

The first term can be developed as:
\begin{equation*}
    \begin{aligned}
        \mathcal{L}_{rec}(\theta,\phi;x_{0:T}) &= -\sum_{t=1}^T  \mathbb{E}_{q_\phi}\left[\log p_\theta(x_t|x_0,z_{1:t})\right] \\
        &= -\sum_{t=1}^T  \mathbb{E}_{q_\phi}\left[\log \mathcal{N}(x_t;\mu_{\theta_x}(z_{1:t}, x_0),Id)\right] \\
        &= \sum_{t=1}^T  \mathbb{E}_{q_\phi}\left[\parallel \mu_{\theta_x}(z_{1:t}, x_0) - x_t \parallel_2^2 \right].
    \end{aligned}
\end{equation*}
This is a reconstruction term between the input and the output of Motion-DVAE.

The second term is a Kullback-Leibler divergence pushing the posterior distribution towards the prior:
\begin{equation*}
    \mathcal{L}_{KL}(\theta,\phi;x_{0:T}) = \sum_{t=1}^T \mathbb{E}_{q_\phi}[D_{KL}(q_\phi(z_t|z_{1:t-1},x_{t:T},x_0) || p_\theta(z_t|z_{1:t-1}, x_0))].
\end{equation*}

We also add a regularization term using the SMPL model to enforce the final human mesh to be as close as possible to the original mesh. It consists of a squared reconstruction error on meshes and joints:
\begin{equation*}
    \mathcal{L}_{reg}(\theta,\phi;x_{0:T}) = \sum_{t=1}^T  \mathbb{E}_{q_\phi}\left[\parallel \mathcal{M}_\beta(\mu_{\theta_x}(z_{1:t}, x_0)) - \mathcal{M}_\beta(x_t) \parallel_2^2 \right].
\end{equation*}

\subsection{Learning settings}\label{dvae_settings}
We train Motion-DVAE with sequences of 30 frames from the AMASS \citesupp{AMASSsupp} dataset, previously downsampled to 30Hz. Then, learning sequences last 1 second. We choose this duration because we argue that even if human motion can last more than 1 second, direct dependencies between poses rarely exceeds 1 second. To ease learning, following \citesupp{rempe2021humorsupp}, we align the first frame of each sequence in the canonical coordinate frame, meaning that translation $r_0$ and the first two components of root-orient $\Phi_0$ are 0. This enables focusing on learning spatial-temporal dependencies independently from the starting point of the motion, which makes the motion prior more general.

We use batches of 64 sequences and train Motion-DVAE for 200 epochs. Similar to HuMoR \citesupp{rempe2021humorsupp}, we use Adamax \citesupp{kingma2014adamsupp} with the same settings and learning rate decays. We also use KL-annealing \citesupp{bowman2015generatingsupp} during the first 50 epochs. However, since our model does not use past predictions for current state prediction, we do not need to perform scheduled sampling.

\section{Unsupervised learned denoising posterior distribution}\label{posterior}
Let's recall the joint distribution in the context of motion denoising:
\begin{equation}\label{jointDistribution}
    p_\theta(y_{0:T},v_{0:T},z_{1:T} | \beta, x_0) = p(v_0)p(y_0|x_0,\beta,v_0) \prod_{t=1}^T p_\theta(z_t|z_{1:t-1},x_0) p(v_t) p_\theta(y_t|z_{1:t},\beta, v_t, x_0).
\end{equation}

We want to express the posterior distribution $p_\theta(x_0,\beta,z_{1:T},v_{0:T}|y_{0:T})$:
\begin{equation*}
    \begin{aligned}
        p_\theta(x_0,\beta,z_{1:T},v_{0:T}|y_{0:T}) &= p(\beta|v_{0:T},z_{1:T},x_0,y_{0:T}) p_\theta(v_{0:T},z_{1:T},x_0|y_{0:T}) \\
        & \\
        &= p(\beta|v_{0:T},z_{1:T},x_0,y_{0:T}) p(v_{0:T}|z_{1:T},x_0,y_{0:T}) p_\theta(z_{1:T}|x_0,y_{0:T}) p(x_0|y_{0:T}) \\
        & \\
        &= p(\beta|v_{0:T},z_{1:T},x_0,y_{0:T}) \prod_{t=1}^T \Bigl[p(v_t|z_{1:t},y_t,x_0) p_\theta(z_t|z_{1:t-1},y_{t:T},x_0)\Bigr] p(v_0|y_0,x_0) p(x_0 | y_{0:T}) \\
        & \\
        &\simeq  p(\beta|y_{0:T}) \prod_{t=1}^T \Bigl[p(v_t|z_{1:t},y_t,x_0) p_\theta(z_t|z_{1:t-1},y_{t:T},x_0)\Bigr] p(v_0|y_0,x_0) p(x_0 | y_{0:T}).
    \end{aligned}
\end{equation*}

As can be seen in the above calculations, we approximate the posterior of $\beta$. We choose to ignore the noise and the latent motion to simplify the model. In practice, $\beta$ is computed as the average of the observed sequence of body shapes.

As a reminder, we define the following approximate posterior:
\begin{equation}
    q(x_0,\beta,z_{1:T},v_{0:T}|y_{0:T})=q(\beta|y_{0:T}) \prod_{t=1}^T \Bigl[q_\gamma(v_t|z_{1:t},y_t,x_0) q_\phi(z_t|z_{1:t-1},y_{t:T},x_0)\Bigr] q_\gamma(v_0|y_0,x_0) q_\omega(x_0 | y_{0:T}).
\end{equation}

\section{ELBO derivation and loss functions}\label{detailed_elbo}
For finetuning Motion-DVAE, we aim to minimize the Kullback-Leibler divergence:
\begin{equation}
    \underset{\phi,\gamma,\omega}{\min} D_{KL}(q_{\phi,\gamma,\omega}(x_0,\beta,z_{1:T},v_{0:T}|y_{0:T}) || p_\theta(x_0,\beta,z_{1:T},v_{0:T}|y_{0:T})).
\end{equation}

We will do it by maximizing the ELBO:
\begin{equation}
    \mathcal{L}(\phi,\gamma,\omega) = \mathbb{E}_{q_{\phi,\gamma,\omega}(x_0,\beta,z_{1:T}, v_{0:T}|y_{0:T})}\left[\log p_\theta(y_{0:T},v_{0:T},z_{1:T},x_0,\beta) - \log q_{\phi,\gamma,\omega}(x_0,\beta,z_{1:T},v_{0:T}|y_{0:T})\right].
\end{equation}

\subsection{ELBO decomposition}
One can notice that the first term of the ELBO involves the joint distribution defined in \cref{jointDistribution}. Taking the logarithm and expectation of this joint distribution, we obtain:
\begin{equation*}
\mathbb{E}_{q_{\phi,\gamma,\omega}}\left[\log p(v_0) + \log p(y_0|x_0,\beta,v_0)\right] + \sum_{t=1}^T  \mathbb{E}_{q_{\phi,\gamma,\omega}}\left[\log p_\theta(z_t|x_0,z_{1:t-1}) + \log p(v_t) + \log p_\theta(y_t|x_0, z_{1:t}, \beta, v_t)\right].
\end{equation*}

Similarly, the second term involves:
\begin{equation*}
    q_{\phi,\gamma,\omega}(x_0,\beta,z_{1:T},v_{0:T}|y_{0:T})=q(\beta|y_{0:T}) \prod_{t=1}^T \Bigl[q_\gamma(v_t|z_{1:t},y_t,x_0) q_\phi(z_t|z_{1:t-1},y_{t:T},x_0)\Bigr] q_\gamma(v_0|y_0,x_0) q_\omega(x_0 | y_{0:T}),
\end{equation*}
which becomes after taking the logarithm and expectation:
\begin{equation*}
    \begin{aligned}
        &\mathbb{E}_{q_{\phi,\gamma,\omega}(x_0,\beta,z_{1:T},v_{0:T}|y_{0:T})}\left[\log q(\beta|y_{0:T}) + \log q_\gamma(v_0|y_0, x_0) + \log q_\omega(y_{0:T}) \right] \\
    &+ \sum_{t=1}^T \mathbb{E}_{q_{\phi,\gamma,\omega}(x_0,\beta,z_{1:T},v_{0:T}|y_{0:T})}\left[\log q_\gamma(v_t|z_{1:t},y_t,x_0) + \log q_\phi(z_t|z_{1:t-1},y_{t:T},x_0)\right].
    \end{aligned}
\end{equation*}

The decomposed ELBO is then:
\begin{equation*}
    \begin{aligned}
        &\mathcal{L}(\phi,\gamma,\omega) = \mathbb{E}_{q_{\phi,\gamma,\omega}(x_0,\beta,z_{1:T},v_{0:T}|y_{0:T})}\left[\log p(y_0|x_0,\beta,v_0) + \sum_{t=1}^T \log p_\theta(y_t|z_{1:t},\beta,v_t,x_0)\right] \\
        &+ \mathbb{E}_{q_{\phi,\gamma,\omega}(x_0,\beta,z_{1:T},v_{0:T}|y_{0:T})}\left[- \log q_\omega(x_0|y_{0:T})\right] \\
        &+ \mathbb{E}_{q_{\phi,\gamma,\omega}(x_0,\beta,z_{1:T},v_{0:T}|y_{0:T})}\left[\sum_{t=1}^T\log p_\theta(z_t|z_{1:t-1},x_0) - \sum_{t=1}^T\log q_\phi(z_t|z_{1:t-1},y_{t:T},x_0)\right] \\
        &+ \mathbb{E}_{q_{\phi,\gamma,\omega}(x_0,\beta,z_{1:T},v_{0:T}|y_{0:T})}\left[\sum_{t=0}^T\log p(v_t) - \sum_{t=1}^T \log q_\gamma(v_t|z_{1:t},y_t,x_0)\right] \\
        &+ \mathbb{E}_{q_{\phi,\gamma,\omega}(x_0,\beta,z_{1:T},v_{0:T}|y_{0:T})}\left[- \log q(\beta|y_{0:T})\right],
    \end{aligned}
\end{equation*}
which can be rewritten as:
\begin{equation*}
    \begin{aligned}
        &\mathcal{L}(\phi,\gamma,\omega) = \mathbb{E}_{q_{\phi,\gamma,\omega}(x_0,\beta,z_{1:T},v_{0:T}|y_{0:T})}\left[\log p(y_0|x_0,\beta,v_0) + \sum_{t=1}^T \log p_\theta(y_t|z_{1:t},\beta,v_t,x_0)\right] \\
        &- \mathbb{E}_{q_{\phi,\gamma,\omega}(x_0,\beta,z_{1:T},v_{0:T}|y_{0:T})}\left[\log q_\omega(x_0|y_{0:T})\right] \\
        &- \mathbb{E}_{q_{\phi,\gamma,\omega}(x_0,\beta,v_{0:T}|z_{1:T},y_{0:T})}\left[D_{KL}(\prod_{t=1}^T q_\phi(z_t|z_{1:t-1},y_{t:T},x_0) \parallel \prod_{t=1}^T p_\theta(z_t|z_{1:t-1},x_0))\right] \\
        &- \mathbb{E}_{q_{\phi,\gamma,\omega}(x_0,\beta,z_{1:T}|v_{0:T},y_{0:T})}\left[D_{KL}(\prod_{t=0}^T q_\gamma(v_t|z_{1:t}, y_t, x_0) \parallel \prod_{t=0}^T p(v_t))\right] \\
        &- \mathbb{E}_{q_{\phi,\gamma,\omega}(x_0,\beta,z_{1:T},v_{0:T}|y_{0:T})}\left[\log q(\beta|y_{0:T})\right].
    \end{aligned}
\end{equation*}

\subsection{ELBO terms calculation}
\subsubsection{Data commitment term}
The first term is a data commitment term. We have:
\begin{equation*}
    \begin{aligned}
        &\mathbb{E}_{q_{\phi,\gamma,\omega}(x_0,\beta,z_{1:T},v_{0:T}|y_{0:T})}\left[\log p(y_0|x_0,\beta,v_0) + \sum_{t=1}^T \log p_\theta(y_t|x_0,z_{1:t},\beta,v_t)\right] \\
        &= \mathbb{E}_{q_{\phi,\gamma,\omega}}\left[\log \mathcal{N}(y_0; \mathcal{M}_\beta(x_0),v_0) + \sum_{t=1}^T \log \mathcal{N}(y_t; \mathcal{M}_\beta(\mu_{\theta_x}(x_0,z_{1:t})),v_t))\right] \\
        &= \frac{1}{2}\mathbb{E}_{q_{\phi,\gamma,\omega}}\Biggl[\sum_{j,d} \frac{1}{v_{0,j,d}} (y_{0,j,d}-\mathcal{M}_\beta(x_0)_{j,d})^2 + \sum_{t,j,d} \frac{1}{v_{t,j,d}}(y_{t,j,d} - \mathcal{M}_\beta(\mu_{\theta_x}(x_0, z_{1:t}))_{j,d})^2\Biggr]
    \end{aligned}
\end{equation*}

As expected, those are Mean Squared Errors weighted by the inverse of the variance.

\subsubsection{Initial state predictor term}
The second term is linked to the initial state predictor:
\begin{equation*}
    \begin{aligned}
        &\mathbb{E}_{q_{\phi,\gamma,\omega}(x_0,\beta,z_{1:T},v_{0:T}|y_{0:T})}\left[\log q_\omega(x_0|y_{0:T})\right] = \mathbb{E}_{q_{\omega}(x_0|y_{0:T})}\left[\log q_\omega(x_0|y_{0:T})\right].
    \end{aligned}
\end{equation*}

This term corresponds to the negative entropy of $q_\omega(x_0|y_{0:T})$. We chose not to use it during learning because it would increase the initial state predicted variance, which is not a desired behavior.

\subsubsection{Motion prior term}
The third term of the ELBO uses Motion-DVAE to implement a motion prior:
\begin{equation*}
    \begin{aligned}
        &\mathbb{E}_{q_{\phi,\gamma,\omega}(x_0,\beta,v_{0:T}|z_{1:T},y_{0:T})}\left[D_{KL}\left(\prod_{t=1}^T q_\phi(z_t|z_{1:t-1},y_{t:T},x_0) \parallel \prod_{t=1}^T p_\theta(z_t|z_{1:t-1},x_0)\right)\right] \\
        &= \sum_{t=1}^T \mathbb{E}_{q_{\omega}q_{\phi}} \Big[ D_{KL}\big(q_\phi(z_t|z_{1:t-1},y_{t:T},x_0) || p_\theta(z_t|x_0,z_{1:t-1})\, \big) \Big] \\
        &= \sum_{t=1}^T \mathbb{E}_{q_{\omega}q_{\phi}} \Big[ D_{KL}\big(\mathcal{N}(z_t;\mu_{\phi}(z_{1:t-1}, x_{t:T}, x_0),\sigma_{\phi}(z_{1:t-1}, x_{t:T}, x_0)) \parallel \mathcal{N}(z_t;\mu_{\theta_z}(z_{1:t-1}, x_0),\sigma_{\theta_z}(z_{1:t-1}, x_0)) \big) \Big] \\
        &= -\frac{1}{2} \sum_{t=1}^T \mathbb{E}_{q_{\omega}q_{\phi}}\left[\log \frac{\sigma_{\mu_{\theta_z}}}{\sigma_\phi} - 1 + \frac{\sigma_\phi}{\sigma_{\mu_{\theta_z}}} + \frac{(\mu_{\mu_{\theta_z}}-\mu_\phi)^2}{\sigma_{\mu_{\theta_z}}}\right].
    \end{aligned}
\end{equation*}

\subsubsection{Noise prior term}
The fourth term of the ELBO is a noise prior term:
\begin{equation*}
    \begin{aligned}
        &\mathbb{E}_{q_{\phi,\gamma,\omega}(x_0,\beta,z_{1:T}|v_{0:T},y_{0:T})}\left[D_{KL}(\prod_{t=0}^T q_\gamma(v_t|z_{1:t}, y_t, x_0) \parallel \prod_{t=0}^T p(v_t)\right] \\
        &= \sum_{t=1}^T \mathbb{E}_{q_{\omega}q_{\phi}}\left[D_{KL}(q_\gamma(v_t|z_{1:t}, y_t, x_0) \parallel p(v_t)\right] \\
        &= \sum_{t,j,d} \mathbb{E}_{q_{\omega}q_{\phi}}\left[ D_{KL}\left(\mathcal{IG}\left(v_{t,j,d},\alpha_\gamma(z_{1:t},y_{t,j,d},x_0),\beta_\gamma(z_{1:t},y_{t,j,d},x_0)\right) \parallel \mathcal{IG}\left(v_{t,j,d},\frac{\lambda}{2},\frac{\lambda}{2}\right)\right)\right] \\
        &= \sum_{t,j,d} \mathbb{E}_{q_{\omega}q_{\phi}}\left[\left(\alpha_\gamma-\frac{\lambda}{2}\right)\psi\left(\alpha_\gamma\right) - \log \Gamma\left(\alpha_\gamma\right) + \log \Gamma\left(\frac{\lambda}{2}\right) + \frac{\lambda}{2}\left(\log \beta_\gamma - \log \frac{\lambda}{2}\right) + \alpha_\gamma \frac{\frac{\lambda}{2} - \beta_\gamma}{\beta_\gamma}\right],
    \end{aligned}
\end{equation*}
where $\Gamma$ and $\psi$ are the Gamma and Digamma functions.

\subsubsection{Body shape term}
The last term is about body shape:
\begin{equation*}
    \begin{aligned}
        &\mathbb{E}_{q_{\phi,\gamma,\omega}(x_0,\beta,z_{1:T},v_{0:T}|y_{0:T})}\left[\log q(\beta|y_{0:T})\right] = \mathbb{E}\left[\delta(\beta-\Bar{\beta}_{SPIN}(y_{0:T}))\right] \\
    \end{aligned}
\end{equation*}

Since that term does not depend on any learned parameter, we ignore it during training.

\section{Final predictions from observations}\label{final_pred}
\subsection{Initial state}
We start by predicting the initial state:
\begin{align}
    \begin{split}
        \hat{x}^{3d}_0 &= \mathbb{E}_{q_{\omega}(x_0|y_{0:T})}[\mathcal{M}_\beta(x_0)] \simeq \mathcal{M}_\beta(\mu_\omega(y_{0:T})),
    \end{split}
\end{align}
where $\mu_\omega$ is the mean vector of $q_\omega(x_0 | y_{0:T})$.

\subsection{Motion prediction}

Then we need to compute $\hat{x}_{1:T} = \mathbb{E}_{p_\theta(x_{1:T}|y_{0:T})}[x_{1:T}]$ where
\begin{equation*}
    \begin{aligned}
        p(x_{1:T}|y_{0:T}) &= \int p_\theta(x_{1:T},x_0,\beta,z_{1:T},v_{0:T}|y_{0:T})\;d_{x_0}d_\beta d_{z_{1:t}} d_{v_{0:T}} \\
        &= \int p_\theta(x_{1:T}|x_0,\beta,z_{1:T},v_{0:T},y_{0:T}) p_\theta(x_0,\beta,z_{1:T},v_{0:T}|y_{0:T})\;d_{x_0}d_\beta d_{z_{1:t}} d_{v_{0:T}} \\
        &= \mathbb{E}_{p_\theta(x_0,\beta,z_{1:T},v_{0:T}|y_{0:T})}\Bigl[p_\theta(x_{1:T}|x_0,\beta,z_{1:T},v_{0:T},y_{0:T})\Bigr].
    \end{aligned}
\end{equation*}
Approximating the posterior $p_\theta(x_0,\beta,z_{1:T},v_{0:T}|y_{0:T})$ with $q_{\phi,\gamma,\omega}$ we obtain:
\begin{align}
\hat{x}_{1:T} = \mathbb{E}_{q_{\phi,\gamma,\omega}}\Bigl[\mathbb{E}_{p_\theta(x_{1:T}|x_0,\beta,z_{1:T},v_{0:T},y_{0:T})}[x_{1:T}]\Bigr].
\end{align}

We need to calculate $p_\theta(x_{1:T}|x_0,\beta,z_{1:T},v_{0:T},y_{0:T})$:
\begin{equation*}
    \begin{aligned}
        \log p_\theta(x_{1:T}|x_0,\beta,z_{1:T},v_{0:T},y_{0:T}) &\overset{c}{=}  \log p_\theta(x_{1:T},x_0,\beta,z_{1:T},v_{0:T},y_{0:T}) \\
        &\overset{c}{=} \sum_{t} \log p(y_t|x_t,v_t)p_\theta(x_t|z_{1:T},x_0) \\
        &= \sum_t \log \mathcal{N}(y_t;x_t,\diag(v_t)) \mathcal{N}(x_t;\mu_{\theta_x},\sigma_{\theta_x}),
    \end{aligned}
\end{equation*}
where $\overset{c}{=}$ denotes equality up to an additive constant that does not depend on $x_{1:T}$. We can further develop the last line and identify:
\begin{equation*}
    \begin{aligned}
        \log p_\theta(x_{1:T}|x_0,\beta,z_{1:T},v_{0:T},y_{0:T})= \sum_{t,j,d} \log \mathcal{N}\left(x_{t,j,d};\frac{v_{t,j,d} (\mu_{\theta_x})_{j,d}+ (\sigma_{\theta_x})_{j,d} y_{t,j,d}}{v_{t,j,d} + (\sigma_{\theta_x})_{j,d}},\frac{v_{t,j,d}(\sigma_{\theta_x})_{j,d}}{v_{t,j,d} + (\sigma_{\theta_x})_{j,d}}\right).
    \end{aligned}
\end{equation*}
This finally leads to:
\begin{equation}
    \hat{x}^{3d}_{t,j,d} = \mathbb{E}_{q_{\phi,\gamma,\omega}}\left[\frac{v_{t,j,d} \mathcal{M}_\beta(\mu_{\theta_x}(x_0,z_{1:t-1})_{j,d})+  y^{3d}_{t,j,d}}{v_{t,j,d} + 1}\right].
\end{equation}

\section{Unsupervised denoising learning}\label{denoising_implementation}
\subsection{Neural networks implementation}
During the unsupervised denoising training, we introduce 2 neural networks: an initial state predictor and a noise predictor. Those 2 models were not necessary for training Motion-DVAE on clean motion capture data since $x_0$ was known and there was no noise in the observations.

As described in the main paper, the initial state predictor implements $q_\omega(x_0|y_{0:T}) = \mathcal{N}(x_0; \mu_\omega(y_{0:T}), \sigma_\omega(y_{0:T}))$. It is implemented by:
\begin{itemize}
    \item $h_0 = 0$
    \item $h_t = r_h(h_{t+1}, y_t)$
    \item $\hat{x_0}=e_{ini}(h_T)$
\end{itemize}

$r_h$ is an anticausal LSTM neural network. We want it to take the observations $y_{0:T}$ backward in time because $x_0$ should depend more on $y_0$ than on $y_T$. $e_{ini}$ is an MLP.

For the noise predictor, we implement $q_\gamma(v_t |z_{1:t}, y_t, x_0) = \prod_{j,d} \mathcal{IG}\left(v_{t,j,d};\alpha_\gamma(z_{1:t},y_t,x_0),\beta_\gamma(z_{1:t},y_t,x_0)\right)$ as follows:
\begin{itemize}
    \item $h_0^z = d_{ini}(x_0)$
    \item $h_t^z = d_{h^z}(h_{t-1}^z, z_t)$
    \item $[\alpha_\gamma, \beta_\gamma] = e_b(h_t,y_t)$
\end{itemize}

$d_{ini}$ and $d_{h^z}$ are the MLP and the LSTM neural networks previously defined for Motion-DVAE. $e_b$ is an MLP only used by the noise predictor.

\subsection{Learning settings}
Unsupervised denoising training for regression mode is performed on training data for 500 epochs, with early stopping when the validation loss does not improve for 10 epochs. Note that for noisy AMASS \citesupp{AMASSsupp} data, the training converges in about 50 epochs only, probably due to a large amount of training data. In optimization mode, we fix the number of iterations (200 iterations on i3DB \citesupp{i3DBsupp}, and 50 for AMASS).

During unsupervised denoising learning, the decoder and prior networks are fixed, preserving the motion prior. We optimize the weights of the encoder, initial state predictor, and noise predictor networks. 
As for training the original Motion-DVAE, we use KL-annealing \citesupp{bowman2015generatingsupp}.

\section{SPIN-t}
\label{sec:spint}
SPIN-t is a procedure built on SPIN's~\citesupp{SPINsupp} frame-wise local pose estimations (in frame coordinates) to obtain global motion (in world coordinates). SPIN provides SMPL pose and shape parameters $\Theta$ and $\beta$, as well as orientation $\phi$ relative to the camera. To obtain motion in global coordinates, we need the global translation $r$. For smooth and realistic 3D trajectories, we decide to optimize both global translation and rotation. We define $v$ as the output of the SMPL model forward passes with optimization variables $r$ and $\phi$ and SPIN~\citesupp{SPINsupp} predictions $\beta$ and $\Theta$. During the optimization, $v$ it only depends on the global translation and rotation. We use L-BFGS~\citesupp{nocedal2006nonlinearsupp} to solve the following optimization problem:
\begin{equation}
    \underset{r,\phi}{\min} \quad \lambda_{data} ~ \xi_{data}(r,\phi) + \lambda_{smooth} ~ \xi_{smooth}(r,\phi),
\end{equation}
where $\xi_{data}(r,\phi) = \underset{t}{\sum} \underset{j}{\sum} \sigma_{t,j} \rho(\Pi(v_{t,j})-y_{t,j})$, with $t$ the temporal index, $j$ the observed joint, $y_{t,j}$ and $\sigma_{t,j}$ the Openpose~\citesupp{openposesupp} 2D detection and its associated confidence, $\rho$ the robust Geman McClure function~\citesupp{geman1987statisticalsupp}, and $\Pi$ the pinhole projection; and $\xi_{smooth} = || v_{1:T} - v_{0:T-1} ||_2^2$. $\xi_{data}$ ensures that the projection of the predicted 3D joints corresponds to the 2D detections, while $\xi_{smooth}$ smooths the motion over time. $\lambda_{data}$ and $\lambda_{smooth}$ are two hyperparameters.

\section{Ablation of SOTA optimization processes}
The optimization process of HuMoR can be divided into 2 parts. A first part called VPoser-t minimizes the projection error of 3D predictions relative to 2D keypoints~\citesupp{openposesupp}, with a pose prior~\citesupp{SMPLifysupp} and regularization terms. The second part of optimization uses a CVAE motion prior to predict more plausible motions. We first decrease the number of iterations of the last part, which is the most time-consuming and the less important for the final results. We then decreased the number of iterations for VPoser-t, leading to significant degradation in performance. 

Pose-NDF uses 5 steps of 50 iterations. At each iteration, the observation reconstruction weight becomes less and less important compared to the pose prior and smoothing term. Then, changing the number of iterations per step would change the balance of the loss functions, and we thus reduce the number of steps instead.

\section{Additional experiments}
\label{sec:additional_exp}

\subsection{Plausibility versus speed}

\begin{figure*}[ht]
    \centering
    {\includegraphics[width=\textwidth]{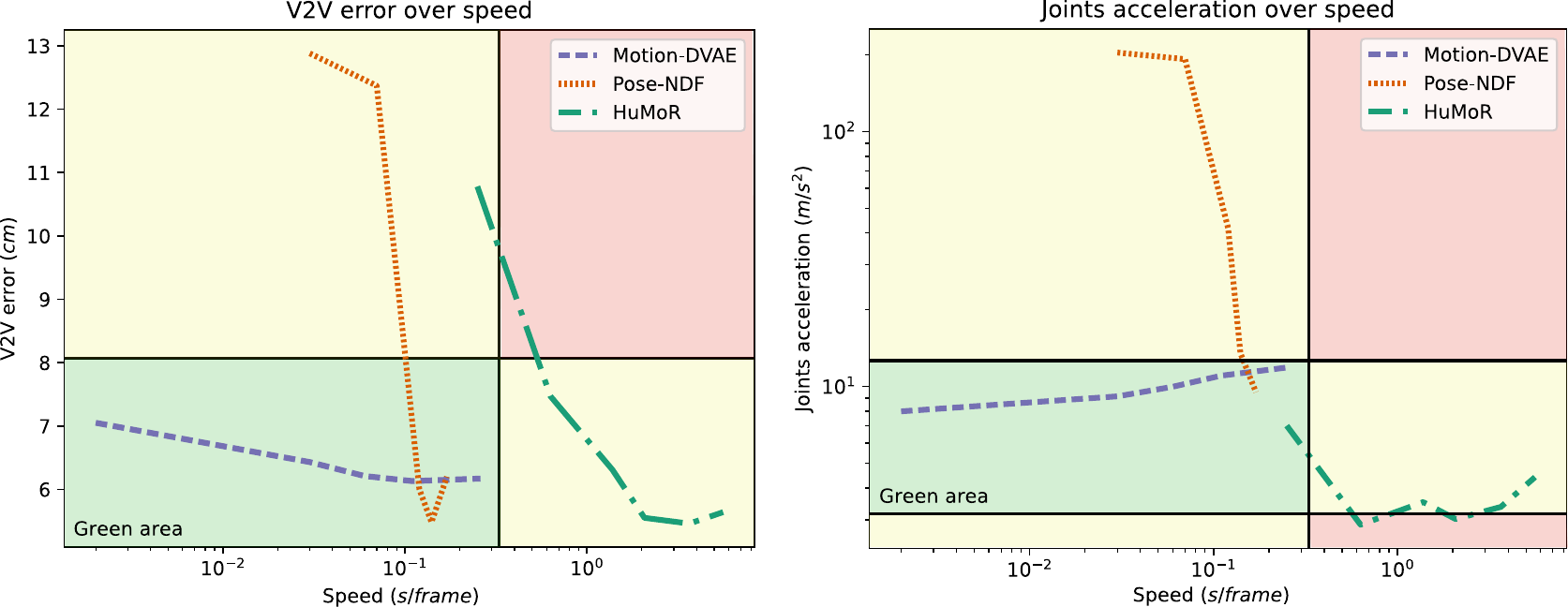}}
    \caption{Evolution of the performance versus speed. Vertex-to-vertex (V2V) error is an accuracy metric, while joints acceleration measures plausibility.}
    \label{plausibility}
\end{figure*}

This experiment is similar to the "Accuracy versus Speed" experiment in the main paper. We consider that joint acceleration should be between half and double of the ground-truth joint acceleration. Results are shown in \cref{plausibility}.

HuMoR obtains satisfying plausibility performance, but it is very slow. Pose-NDF produces motions that are not smooth enough. Considering plausibility and speed Motion-DVAE is clearly the best choice. We observe that the more optimization iterations are performed with the proposed method, the better accuracy, but also the less smooth predictions. This result is not surprising since optimizing the model for specific samples leads to a loss of generalization. 

\subsection{Experiments on i3DB}

\begin{figure*}[ht]
    \centering
    {\includegraphics[width=\textwidth]{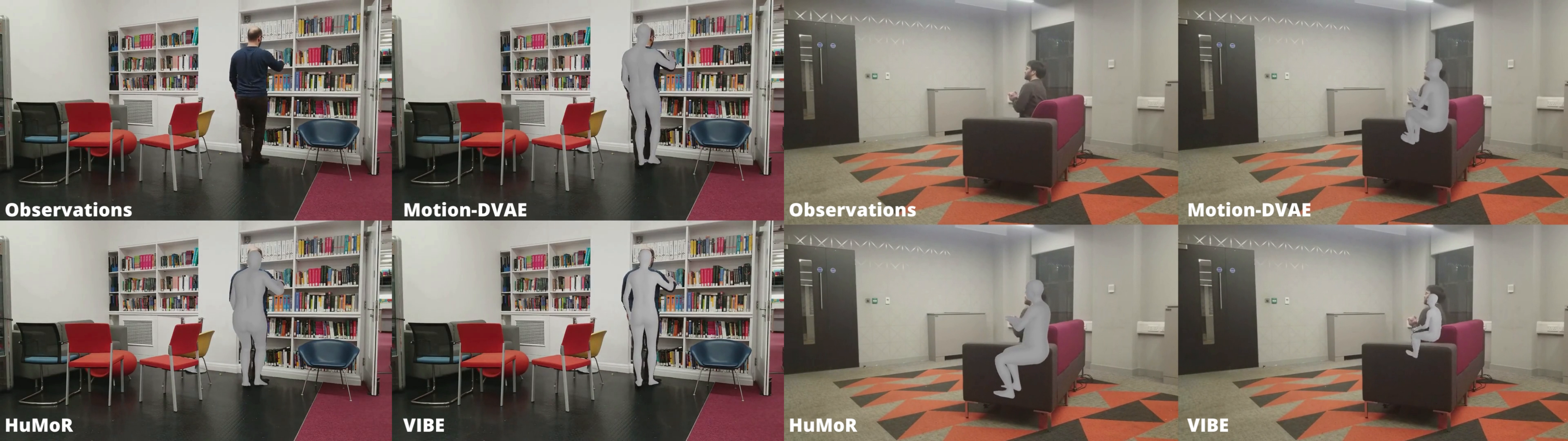}}
    \caption{Qualitative results for motion estimation from videos.}
    \label{qualRGB}
\end{figure*}

We provide supplementary experiments on i3DB~\citesupp{i3DBsupp}. The i3DB dataset~\citesupp{i3DBsupp} contains in-the-wild videos with medium to heavy occlusions due to person-environment interaction. It provides global 3D annotations on joint coordinates that we use for computing evaluation metrics. We initialize our method using an off-the-shelf 2D pose estimation~\citesupp{openposesupp} and SPIN~\citesupp{SPINsupp}, a method providing frame-wise SMPL pose estimates.

\subsubsection{Evaluation on RGB videos}
We evaluate the proposed method on i3DB~\citesupp{i3DBsupp}. Following~\citesupp{rempe2021humorsupp}, we only use six scenes with accurate annotations (Scenes 05, 07, 10, 11, 13, and 14). Due to this dataset's low number of videos, we evaluate our approach in a cross-validation setting, with each fold: 4 videos for training, 1 for validation, and 1 for testing. As in the previous experiment, the method is evaluated in the two regression and optimization modes.

As for the motion denoising experiment, we use very high values for noise parameter $\lambda$ for learning the denoising process on the noisy training set. However, we use a robust learned noise model in the optimization mode. This gives a Student-t prior distribution for the noise on the 3D observations. Fitting a robust noise model on noisy observation sequences allows us to adapt to the observed noise in an unsupervised manner, which is expected to be beneficial in terms of prediction accuracy. Indeed, as explained in the section 4.3 of the main paper, this unsupervised adaptation allows us to find a compromise between the noisy observations and the Motion-DVAE output, depending on the estimated noise variance. A lower variance will give more importance to the observations and vice versa.

We use the open-source implementations of SPIN~\citesupp{SPINsupp}, VIBE~\citesupp{kocabas2019vibesupp}, and HuMoR~\citesupp{rempe2021humorsupp} for comparing with SOTA algorithms. We also include MVAE~\citesupp{ling2020charactersupp} in the comparison, but taking the results given in the paper of~\citesupp{rempe2021humorsupp}. We do not have the running time for this method, but we can assume that it is similar to HuMoR since MVAE can be thought of as ablation of the HuMoR CVAE using the same optimization procedure.
The running time of 2D keypoints detection by off-the-shelf models is not considered in speed calculation since it is used by all methods. However, we add SPIN execution time to our method for a fair comparison with SPIN~\citesupp{SPINsupp} and other methods. Results are presented in \cref{i3dbQuant}.

\begin{table*}[ht]
    \centering
    \caption{Results on i3DB: We compare the proposed method (bottom part) and SOTA (top part) performance in terms of global and local Mean Per Joint Positional Error (MPJPE) (cm) and execution time (sec/frame) on visible (Vis) and occluded (Occ) joints. We achieve an accuracy competitive with the SOTA methods while showing reasonably high speed.}
    {\begin{tabular}{c|c|ccc|ccc}
                                & {Speed $\downarrow$} & \multicolumn{3}{c|}{G-MPJPE $\downarrow$}   & \multicolumn{3}{c}{MPJPE $\downarrow$}                       \\
                                & (sec/frame)    & All              & Vis               & Occ               & All               & Vis               & Occ                   \\
\midrule
SPIN~\citesupp{SPINsupp}                & \underline{0.05} & -              & -                 & -                 & 14.95             & 11.89             & 23.90                  \\
VIBE~\citesupp{kocabas2019vibesupp}     & \textbf{0.02}  & 116.46           & 90.05             & 192.55            & 15.08             & 12.06             & 23.78                 \\
MVAE~\citesupp{ling2020charactersupp}   & -              & 40.91            & 37.54             & 50.63             & 19.17             & 16.00             & 28.32                 \\
VPoser-t~\citesupp{rempe2021humorsupp}  & 1.05          & \underline{31.88} & \underline{28.84} & \underline{40.63} & 16.36             & 12.81             & 26.57                 \\
HuMoR~\citesupp{rempe2021humorsupp}     & 11.37          & \textbf{28.68}   & \textbf{26.01}    & \textbf{36.37}    & 15.29             & 12.57             & 23.10                 \\
\hline
SPIN-t (ours)                   & 0.25           & 42.45            & 36.54             & 59.71             & 14.28             & 11.66             & 21.94                 \\
Regression (ours)               & 0.26           & 41.41            & 36.55             & 55.45             & \textbf{13.82}    & \underline{11.54} & \textbf{20.40}        \\
Optimization (ours)             & 0.56           & 41.09            & 36.01             & 55.78             & \underline{14.08} & \textbf{11.48}    & \underline{21.58}                \end{tabular}}
    \label{i3dbQuant}
\end{table*}

We can notice that the proposed SPIN-t improves the SPIN predictions regarding the MPJPE by adjusting the global rotation. SPIN-t outperforms all SOTA methods in terms of MPJPE. However, VPoser-t is more accurate globally (i.e., in terms of G-MPJPE) by about 8~cm.

Motion-DVAE outperforms all SOTA models by more than 1cm in terms of MPJPE. Regarding the global error, Motion-DVAE is similar to MVAE and significantly outperforms VIBE, which gives inaccurate results for occluded joints. HuMoR obtains better global predictions but is prohibitively slow for real-world applications.

Qualitative results are shown in \cref{qualRGB}. When there are no occlusions, all methods perform great visually. However, with occlusions, there are clear differences between predictions. VIBE fails to estimate the global translation, and the arms pose is not as accurate as other methods. Motion-DVAE and HuMoR give good local predictions, with more realistic floor contacts for the latter. HuMoR seems to underestimate the distance between the person and the camera for global predictions as the prediction looks closer than the person on the image, while Motion-DVAE predicted translation is a bit too far relative to the camera.
Qualitative results for videos are made available in the supplementary material.

\subsubsection{Ablation study}
We also perform an ablation study to evaluate the impact of the different components of the denoising procedure learned in an unsupervised fashion. We chose i3DB~\citesupp{i3DBsupp} for the ablation study because this dataset has many occlusions, enabling us to evaluate the model components for dealing with occlusions. Results are shown in \cref{ablationQuant}. Most ablations are tested in regression (Reg) and optimization modes (Opt), except for the last experiment, which only concerns the optimization mode. Results are compared with the performance without any ablation (Full).

The first ablation experiment, "NN output", predicts the initial state, the latent motion, and the noise by taking directly the expectations outputted by the associated inference model networks successively instead of sampling the approximate posterior distributions as explained in the section 4.3 of the main paper. This approximation is made in most works, but neural network decoders are not linear functions. Thus taking the expectation of the latent representation and decoding it does not give the expectation of the network output. Results obtained without the proposed inference scheme are less accurate by about 1.5~cm in the regression mode and 0.5~cm in the optimization mode.

Next, we remove $\mathcal{L}_{KL}^{DVAE}(\phi,\omega)$ from the equation (13) of the main paper for both training and optimization ("No prior"). We compare our results with HuMoR~\citesupp{rempe2021humorsupp} by setting the motion prior weight to 0 in optimization mode. Surprisingly, when removing the motion prior, HuMoR obtains better results for occluded and visible joints. On the contrary, removing the motion prior penalizes our method in regression and optimization modes. In the optimization mode, the results get better on every metric using the motion prior. This shows that the proposed motion prior is efficient for unsupervised optimization.

In "No noisy train", we skip the unsupervised training step on noisy data: we perform direct inference and optimize from Motion-DVAE learned on AMASS. As expected, the results deteriorate significantly in the regression mode because we use a model trained on clean data for processing noisy observations. In optimization mode, we do not reach the accuracy of the regression mode with no ablation. This demonstrates the efficiency of the introduced unsupervised learning procedure for denoising new unknown motions.

Finally, in "Gaussian noise" we test the proposed approach in optimization mode using a Gaussian noise prior distribution instead of Student-t. The results are improved on visible joints but deteriorate on occlusions. However, the results in the regression mode are still better than the results in the optimization mode, which shows the excellent generalization of the unsupervised learning approach.

\begin{table}[]
    \centering
    \caption{Ablation study: MPJPE and G-MPJPE are in cm.}
    {\begin{tabular}{c|c|cc|cc}
                    &                               & \multicolumn{2}{c|}{G-MPJPE}  & \multicolumn{2}{c}{MPJPE} \\
  \midrule
                    &                               & Vis   & Occ           & Vis   & Occ       \\
Full                & HuMoR~\citesupp{rempe2021humorsupp}   & 26.01 & 36.37         & 12.57 & 23.10     \\
                    & Regression                    & 36.55 & 55.45         & 11.54 & 20.40     \\
                    & Optimization                  & 36.01 & 55.78         & 11.48 & 21.58     \\
\hline
NN output           & Regression                    & 39.54 & 56.64         & 13.69 & 22.32     \\
                    & Optimization                  & 36.42 & 55.33         & 11.80 & 21.95     \\
\hline
No prior            & HuMoR~\citesupp{rempe2021humorsupp}   & 25.64 & 36.12         & 12.13 & 22.58     \\
                    & Regression                    & 37.51 & 58.77         & 11.64 & 20.90     \\
                    & Optimization                  & 36.75 & 58.68         & 11.56 & 21.74     \\
\hline
No noisy train      & Regression                    & 68.19 & 81.72         & 32.30 & 47.56     \\
                    & Optimization                  & 39.97 & 57.64         & 12.26 & 22.49     \\
\hline
Gaussian noise      & Optimization                  & 35.84 & 56.93         & 11.52 & 21.76
\end{tabular}}
    \label{ablationQuant}
\end{table}